\documentclass[10pt,twocolumn,letterpaper]{article}

\usepackage[pagenumbers]{cvpr} 
\usepackage{multirow}
\usepackage{float}

\usepackage[dvipsnames]{xcolor}

\definecolor{cvprblue}{rgb}{0.21,0.49,0.74}
\usepackage[pagebackref,breaklinks,colorlinks,citecolor=cvprblue]{hyperref}

\def\paperID{2811} 
\def\confName{CVPR}
\def\confYear{2024}

\title{Bootstrapping Chest CT Image Understanding by Distilling Knowledge from X-ray Expert Models}

\author{Weiwei Cao$^{1,2,3}$ \quad Jianpeng Zhang$^{1,5,6}$\thanks{Correspondence to Jianpeng Zhang and Jian Zheng. The work was done during Weiwei Cao's internship at DAMO Academy.} \quad Yingda Xia$^{1}$ \quad Tony C. W. Mok$^{1,6}$ \quad Zi Li$^{1,6}$ \\ Xianghua Ye$^{4}$ \quad Le Lu$^{1}$ \quad Jian Zheng$^{2,3 *}$ \quad Yuxing Tang$^{1}$ \quad Ling Zhang$^{1}$\\
\normalsize{\textsuperscript{1}DAMO Academy, Alibaba Group} \quad
\normalsize{\textsuperscript{2}University of Science and Technology of China, China}\\
\normalsize{\textsuperscript{3}Suzhou Institute of Biomedical Engineering and Technology, Chinese Academy of Sciences, China}\\
\normalsize{\textsuperscript{4}The First Affiliated Hospital of College of Medicine, Zhejiang University, China}\\
\normalsize{\textsuperscript{5}College of Computer Science and Technology, Zhejiang University, China} \quad
\normalsize{\textsuperscript{6}Hupan Lab, 310023, Hangzhou, China}\\
{\tt\small {jianpeng.zhang0@gmail.com}}
}

\begin{document}
\maketitle
\begin{abstract}
Radiologists highly desire fully automated versatile AI for medical imaging interpretation. However, the lack of extensively annotated large-scale multi-disease datasets has hindered the achievement of this goal.
In this paper, 
we explore the feasibility of leveraging language as a naturally high-quality supervision for chest CT imaging.
In light of the limited availability of image-report pairs, we bootstrap the understanding of 3D chest CT images by distilling chest-related diagnostic knowledge from an extensively pre-trained 2D X-ray expert model. 
Specifically, we propose a language-guided retrieval method to match each 3D CT image with its semantically closest 2D X-ray image, and perform pair-wise and semantic relation knowledge distillation. 
Subsequently, we use contrastive learning to align images and reports within the same patient while distinguishing them from the other patients. However, the challenge arises when patients have similar semantic diagnoses, such as healthy patients, potentially confusing if treated as negatives. We introduce a robust contrastive learning that identifies and corrects these false negatives. 
We train our model with over 12K pairs of chest CT images and radiology reports. 
Extensive experiments across multiple scenarios, including zero-shot learning, report generation, and fine-tuning processes, demonstrate the model’s feasibility in interpreting chest CT images.
\end{abstract}    
\section{Introduction}
\label{sec:intro}

\begin{figure}
    \centering
    \includegraphics[width=1\linewidth]{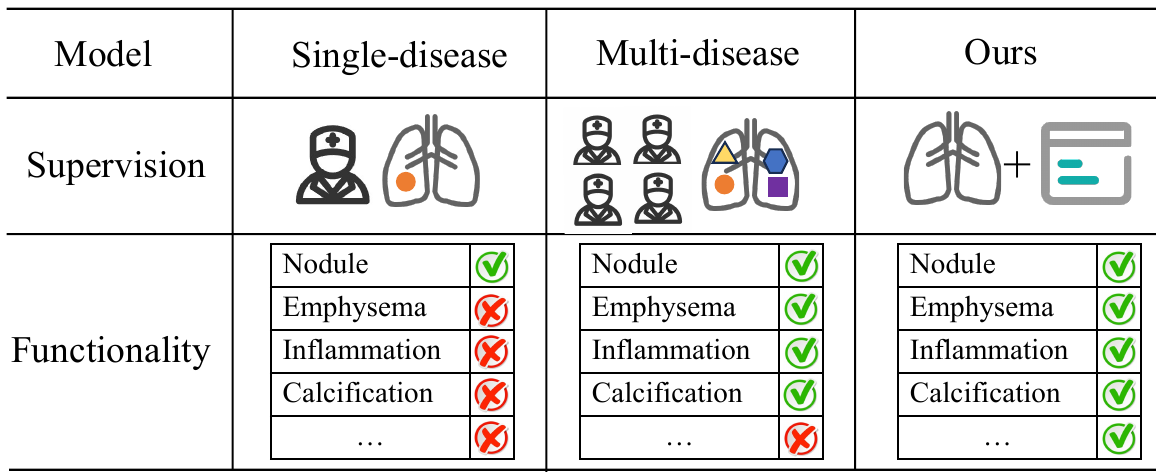}
    \vspace{-0.6cm}
    \caption{Models tailored for specific diseases demand doctors to annotate each image. Creating a multi-disease model involves more time and effort for comprehensive data annotation. In contrast, our model learns to diagnose various diseases from both images and reports, eliminating the need for additional annotations.}
    \vspace{-0.4cm}
    \label{fig:fig1}
\end{figure}

Understanding medical images is crucial for precise clinical diagnosis, with doctors routinely engaging in multi-disease detection and diagnosis in imaging scans. Deep learning models require extensive datasets with fully annotated diseases for comprehensive training~\cite{lecun2015deep, shen2017deep, litjens2017survey}.
In practice, the prevailing approach to developing a diagnostic model typically concentrates on a specific disease, as illustrated in Figure~\ref{fig:fig1}. 
This workflow results in models with limited usage, failing to be applied to other diseases. To augment the capability for a new disease, a significant additional effort is needed for data collection, annotation, and model training, which is not scalable, as there might be hundreds of diseases and tasks. 
Therefore, there is a strong anticipation for a new model development paradigm beyond the current labor-intensive annotation and task-specific training. Imaging diagnostic reports, the core output of radiologists' intellectual work, 
inherently offer disease-level labels. 
The challenge lies in effectively leveraging this text data to enhance medical image understanding.

Recent advances in vision-language pretraining (VLP) have yielded valuable insights~\cite{CLIP2021, BLIP, bao2022vlmo, li2021align, gao2022pyramidclip}. Contrastive learning, a powerful technique in VLP, effectively bridges the image and text modalities. This method unifies their representations in a semantic space by attracting positive image-text pairs while separating negative pairs. VLP models, trained on extensive paired datasets, demonstrate impressive proficiency in image understanding, including capabilities like zero-shot learning~\cite{CLIP2021,han2021contrastive,xu2021videoclip}. 
This progress has notably impacted the medical imaging field, particularly in chest X-ray zero-shot diagnosis~\cite{ChestZero2022, cheng2023prior, huang2021gloria, boecking2022making}. Applying contrastive learning on datasets like MIMIC-CXR~\cite{MIMIC_dataset}, with over 300,000 image-report pairs, has led to diagnostic accuracy comparable to medical professionals in certain lung conditions~\cite{ChestZero2022}. However, it remains challenging to extend this success to the 3D imaging domain, due to limited extensive 3D image volumes and report data.

In this paper, we propose to \textbf{b}ootstrap chest CT \textbf{i}mage \textbf{u}nderstanding by \textbf{d}istilling (BIUD) lung-related medical knowledge from pre-trained chest X-ray expert models. 
Our choice of distilling knowledge from the X-ray expert model is motivated by its outstanding capabilities, facilitated by the abundance of publicly available paired data. Additionally, chest X-rays and CT scans share the same medical knowledge base, especially in lung diseases, making them ideal for cross-referencing. 
There are undeniably visual distinctions between CT and X-ray images, with CT offering more detailed anatomical information. Nevertheless, the X-ray is sufficient for the initial diagnosis of major diseases. Leveraging pathology as a bridge, we aim to equip the CT model with an understanding of the distinctions and similarities in pathology presentation between CT and X-ray, thus enriching its pathology detection capabilities.
The lack of paired CT and X-ray datasets, however, complicates direct knowledge transmission. To overcome this challenge, we propose a language-guided retrieval strategy that empowers CT models to identify semantically aligned images from the extensive X-ray database, thereby facilitating knowledge distillation. This method efficiently conveys expertise from the X-ray model to the CT model in the absence of paired data, enhancing their semantic interpretation abilities. We also introduce a novel robust contrastive learning method (RoCo) for precise semantic alignment of CT image volumes and reports. This is crucial since many reports share semantic similarities, particularly those for healthy patients. Dominant contrastive learning models often misclassified these as false negatives, relying on pair-matching rules. The proposed RoCo objective proactively corrects these misclassifications, ensuring the semantic integrity of the alignment process. Our BIUD model, trained on over 12,000 pairs of chest CT volumes and reports, underwent rigorous validation across diverse scenarios and datasets, including zero-shot learning, report generation, and fine-tuning. It consistently demonstrated superior performance compared to other methods. Notably, a reader study indicated that our model's zero-shot diagnostic capability rivals that of radiologists in specific tasks, markedly reducing the time and resource expenditure.
To summarise, our contributions are three-fold: 
\begin{itemize}
  \item We propose to enhance chest CT image understanding by distilling knowledge from a pre-trained chest X-ray expert model to the CT model. A language-guided retrieval approach is used to match the semantically similar X-ray image for each CT volume, aiming at facilitating knowledge distillation.
  \item We improve the semantic alignment in contrastive learning by identifying and correcting false negative pairs, and direct the model's focus toward crucial entities and attributes by using an entity-focused masking strategy. 
  \item Our BIUD model outperforms existing VLP competitors in zero-shot classification and report generation tasks, evaluated for disease diagnosis across multiple datasets. Importantly, we are the first to demonstrate that the VLP-based method can achieve performance similar to that of an experienced radiologist in diagnosing certain primary diseases in 3D CT imaging. 

\end{itemize}
\section{Related work}
\label{sec:related_work}

\begin{figure*}
    \centering
    \includegraphics[width=0.9\linewidth]{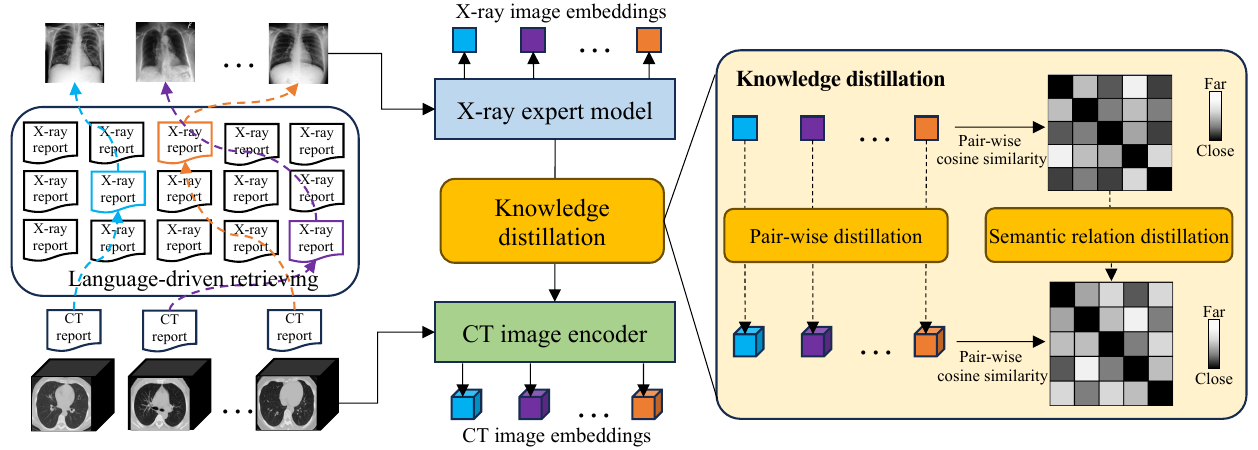}
    \vspace{-0.4cm}
    \caption{Illustration of distilling knowledge from the X-ray expert model to the CT image encoder.}
    \vspace{-0.6cm}
    \label{fig:distillation}
\end{figure*}

\subsection{Language-driven image understanding}
Language serves as an advanced medium for human thought and a valuable source of high-quality supervision in machine learning. Language-driven image understanding endeavors to enable machines to interpret images effectively by utilizing the context provided by natural language. The principal challenge in this domain is achieving semantic alignment between different modalities~\cite{yang2022vision,ma2023liv,wei2023learning, singh2022flava, li2022fine, xie2023medim}. 
CLIP~\cite{CLIP2021}, a notable model introduced in 2021, employs contrastive learning to align images with corresponding textual descriptions in a unified embedding space. By training on extensive datasets of image-text pairs, CLIP simultaneously learns visual and textual information, demonstrating impressive zero-shot reasoning abilities.
In medical imaging, language-driven understanding leverages textual reports to enhance the interpretation of radiological images. For instance, Zhang et al.~\cite{ConVIRT2022} introduced a contrastive learning framework that uses text descriptions to learn visual representations from medical images, thereby improving medical image comprehension. Tiu et al.~ \cite{ChestZero2022} showed that a language-supervised model, trained on chest X-ray images without explicit annotations, can achieve pathology classification accuracy on par with radiologists. This model also generalizes to various tasks, including identifying unseen pathologies, accommodating multiple interpretations, and adapting to datasets from different institutions.
Recent efforts in the medical field have included incorporating domain knowledge through knowledge graphs. Zhang et al.~\cite{KDA2023} proposed the knowledge-enhanced approach, which uses existing medical domain knowledge to guide vision-language pre-training with chest X-rays and radiology reports. Li et al.~ \cite{DCL} introduced a dynamic knowledge graph with adaptable structure and nodes, aimed at improving X-ray image and report alignment.

However, most research has concentrated on 2D chest X-ray scenarios~\cite{ConVIRT2022, ChestZero2022, KDA2023, MedCLIP2022}, with limited exploration in 3D imaging. This is primarily due to the abundance of publicly available image-text datasets for 2D chest X-rays~\cite{MIMIC_dataset, johnson2019mimic, demner2016preparing}, in contrast to the scarcity of 3D imaging resources. 3D imaging presents additional challenges with its complex anatomical structures and detailed report descriptions, complicating the task of image-text alignment. This paper, therefore, focuses on the alignment of 3D medical images and reports, exploring new frontiers in language-driven image understanding in the 3D medical domain.

\subsection{Knowledge distillation}
Knowledge distillation~\cite{hinton2015distilling,gou2021knowledge} aims to compress a large-parameter teacher model into a compact student model while maintaining comparable performance. Recent advancements, such as SAKD~\cite{SAKD}, SPKD~\cite{SPKD}, relational KD~\cite{relationalKD}, and attention transform~\cite{zagoruyko2016paying}, have demonstrated impressive results by learning semantic representations and correlations from teacher models. In the medical imaging field, knowledge distillation has been applied to transfer knowledge from multimodal to unimodal models. These approaches have shown promising outcomes, especially in scenarios involving missing modalities~\cite{multimodalKD, wang2023prototype} and semi-supervised learning~\cite{semiKD}.
%
Different from previous work, our distillation scenario is challenging due to the absence of paired CT and X-ray, which makes cross-modal knowledge transfer difficult. To overcome this challenge, we utilize the reports as a bridge to identify semantically matched pair data, enabling us to achieve the goal of distillation.

\section{Approach}

\subsection{Distilling knowledge from X-ray expert model}

\noindent \textbf{X-ray expert model: }
The distillation aims to transfer the domain knowledge from the teacher model to the student model. Considering our target task is chest CT image understanding, we have opted for the CheXzero model~\cite{ChestZero2022}, pre-trained on MIMIC-CXR, as the expert (teacher). This choice is informed by several key factors. Firstly, MIMIC-CXR offers an extensive repository of publicly available chest X-ray image-report pairs, providing a necessary foundation for training expert models with competencies comparable to medical professionals~\cite{ChestZero2022,KDA2023}. Secondly, the chest X-ray reports of MIMIC-CXR present a rich source of knowledge on pulmonary abnormalities and diseases, encompassing 14 distinct lung conditions. 
CheXzero comprises two primary components: an image encoder, denoted as $F_{img}^{XR}$, and a text encoder, $F_{text}^{XR}$. It is essential to recognize that this image encoder excels in extracting embedding specifically for X-ray images. However, this capability presents a challenge in our scenario: the lack of matched X-ray images corresponding to CT data obstructs the straightforward application of this model to CT images. 
To address this issue, we propose a language-driven retrieval method that singles out the mostly matched X-ray image within the MIMIC-CXR dataset for each CT image. The retrieval process is steered by the level of similarity between their respective reports to guarantee an accurate and contextually relevant pairing.

\noindent \textbf{Language-guided retrieval: }
We denote the X-ray images and reports from the MIMIC-CXR dataset as $I^{XR}=\{I_1^{XR}, I_2^{XR}, ..., I_n^{XR}\}$ and $R^{XR}=\{R_1^{XR}, R_2^{XR}, ..., R_n^{XR}\}$, respectively. Similarly, the CT images and reports are represented as $I^{CT}=\{I_1^{CT}, I_2^{CT}, ..., I_m^{CT}\}$ and $R^{CT}=\{R_1^{CT}, R_2^{CT}, ..., R_m^{CT}\}$, where $n>>m$.
We extract both the X-ray and CT report representations by using the CheXzero model: $f^{XR} = F_{text}^{XR} (R^{XR}) \in \mathbb{R}^{n \times d1}$ and $f^{CT} = F_{text}^{XR} (R^{CT}) \in \mathbb{R}^{n \times d1}$, where $d1$ represents the dimension of text embedding. 
It is important to note that the CT report embeddings are extracted using the expert text encoder to maintain semantic consistency with the X-ray reports.  
The corresponding set of X-ray images retrieved for CT images is derived as
$I^{*}=\{I_1^{*}, I_2^{*}, ..., I_m^{*}\}$, where $I_i^{*} = I_j^{XR} $ and $j = {\arg\max \frac{f_i^{CT}\cdot f_j^{XR}}{\left\| f_i^{CT} \right\| \left\| f_j^{XR} \right\|}}$. 
By utilizing the most similar reports as cues, we are able to identify the X-ray image that bears the closest semantic resemblance to each CT query image.

\noindent \textbf{Knowledge distillation: }
The retrieval procedure identifies pairs of semantically similar CT and X-ray images, $I^{CT}$ and $I^{*}$, which serve as a vital link for facilitating the process of knowledge distillation. As shown in Figure~\ref{fig:distillation}, we extract their respective representations using the CT image encoder and the X-ray expert model, $h^{CT} = F_{img}^{CT} (I^{CT}) \in \mathbb{R}^{m \times d2}$, $h^{*} = F_{img}^{XR} (I^{*}) \in \mathbb{R}^{m \times d2}$, where $d2$ is the dimension of image embedding.
We introduce a dual distillation method, composing of pair-wise distillation and semantic relation distillation. 
Pair-wise distillation brings the representation of CT images closer to their corresponding X-ray representations. 
Furthermore, inspired by \cite{relationalKD, SAKD, SPKD}, we utilize the semantic relation learned from the X-ray expert model as guidance to refine the semantic relation between CT images, aiming at improving the semantic discrimination ability of the CT image encoder. 
Specifically, we calculate the cosine similarity between pairwise X-ray images, resulting in a relation matrix $p^{*} \in \mathbb{R}^{m \times m}$. Similarly, we compute the relation matrix $p^{CT} \in \mathbb{R}^{m \times m}$ for CT images. 
The mean square error (MSE) loss is used to maximize the representational similarity between the CT and X-ray models, expressed as:
\begin{equation}
    L_{dist} = \left\| h^{CT} - h^{*} \right\|_2^2 + \left\| p^{CT} - p^{*} \right\|_2^2.
\end{equation}
Through dual distillation strategies, the CT image encoder is able to assimilate semantic knowledge about chest abnormalities and diseases from the X-ray expert model. This is achieved even without extensive pre-training on CT-report pair data, thereby bolstering the encoder's capacity for semantic alignment.


\subsection{Image-report alignment}
\begin{figure}
    \centering
    \includegraphics[width=1.0\linewidth]{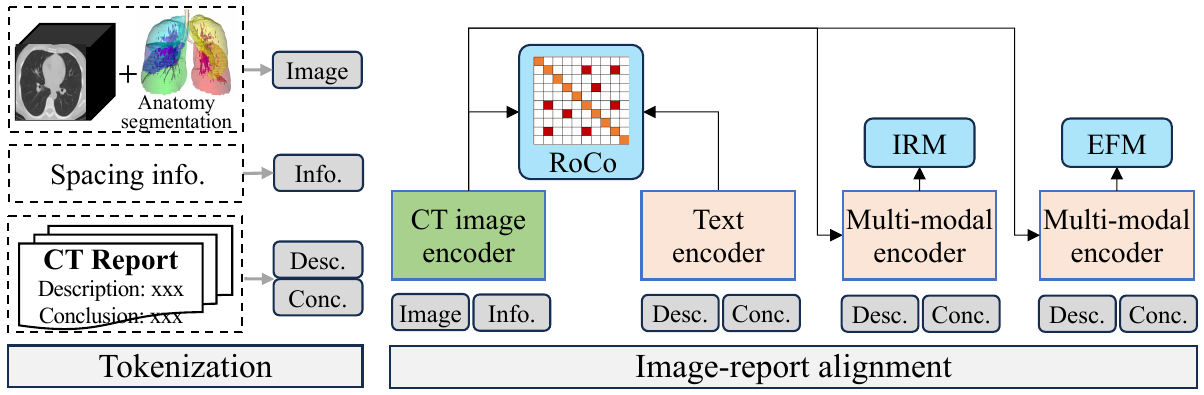}
    \vspace{-0.6cm}
    \caption{Framework of CT image-report alignment.}
    \vspace{-0.4cm}
    \label{fig:architecture}
\end{figure}

Figure~\ref{fig:architecture} illustrates the core modules of image-report alignment, which include a CT image encoder, a text encoder, and a multi-modal encoder. As all modules are built upon the Transformer architecture, data from different modalities are transformed into sequences through tokenization before being fed into the respective modules.
The CT image encoder and text encoder extract embeddings from CT images and reports, respectively. These embeddings are then semantically aligned using a robust contrastive learning (RoCo) approach, elaborately detailed in Sec.~\ref{sec.RoCo}. The multi-modal encoder utilizes cross-attention to fuse report and CT image features for image-report matching (IRM). Additionally, we introduce an entity-focused masking (EFM) strategy to enhance the recognition of keywords, such as entities and attributes, which is comprehensively detailed in Sec.~\ref{sec.EFM}.
In terms of network architecture, the image encoder employs a 12-layer ViT/B, where the original 2D embedding layer of ViT is replaced by a 3D layer to accommodate 3D images. Both the text encoder and the multi-modal encoder utilize a 12-layer Transformer architecture pre-trained with BERT. Notably, these two encoders share parameters, enabling efficient training while capitalizing on the advantages of multi-task learning.

\subsubsection{Incorporating anatomical and physical prior}
To enhance the model's ability to locate and perceive the size of findings, we incorporate the prior information of anatomical structure and voxel spacing into the model.
We utilize a segmentation model to parse the anatomical structures within the lung region, including the left upper lobe, left lower lobe, right upper lobe, right middle lobe, right lower lobe, vessel, and trachea. This information aids the model in accurately locating the specific positions of findings. The image and structure segmentation mask are combined through concatenation and then split into tokens as done in ~\cite{dosovitskiy2020image}. 
In addition to the positional information, the model should possess accurate assessment capabilities for the size of findings, such as ``nodule size is 5mm$\times$ 6mm" mentioned in the report. However, the model is unaware of the true physical spacing represented by a voxel. 
Therefore, it's essential to supply this spacing information to the image encoder, enabling it to generate representations that take into account the physical size of the image. 
We deploy an embedding layer specifically designed to encode voxel size in the (x, y, z) directions into three separate tokens representing spacing information. Subsequently, these tokens are concatenated with the image patch tokens, creating a unified input stream for the image encoder.

\subsubsection{Reducing false negatives in contrastive learning}
\label{sec.RoCo}
\begin{figure}
    \centering
    \includegraphics[width=1\linewidth]{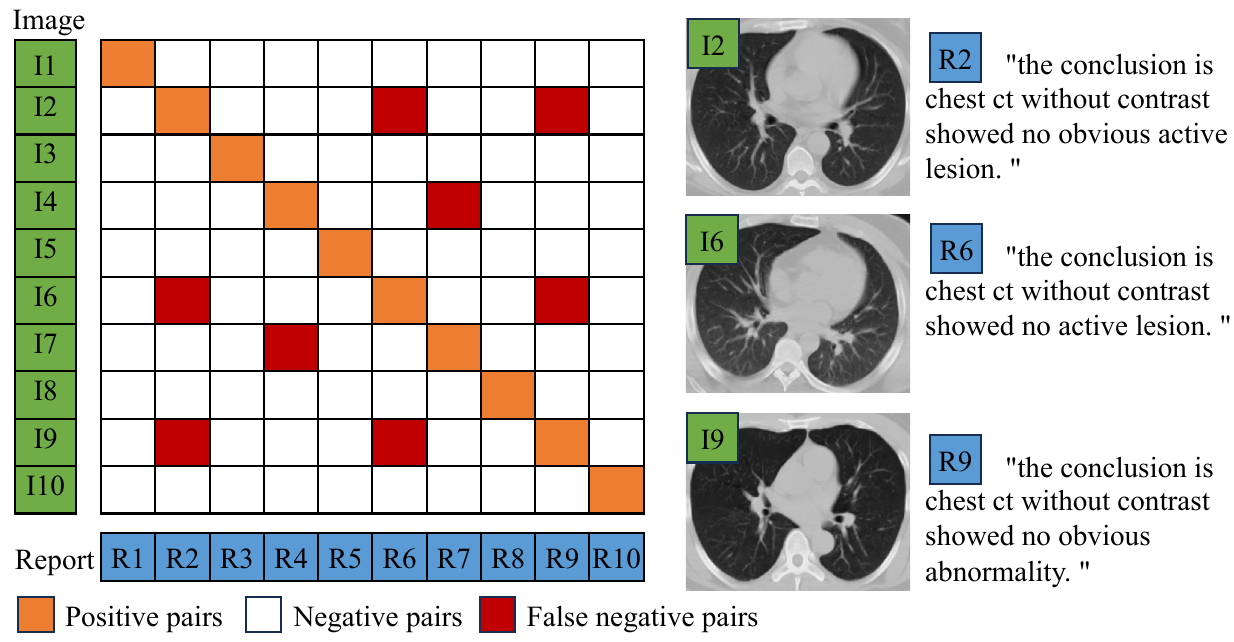}
    \vspace{-0.4cm}
    \caption{The left illustrates the positive pairs, negative pairs, and false negative pairs in robust contrastive learning. The right shows the images and reports of the second, sixth, and ninth samples. These reports all indicate that the patient's lungs are healthy and without abnormalities.}
    \vspace{-0.4cm}
    \label{fig:Pos_FN_pairs_vis}
\end{figure}

Given a mini-batch containing $N$ image-report pairs $\{I, R\}$, the essence of contrastive learning lies in narrowing the distance between image $I_i$  and its matched report $R_i$, while simultaneously distancing the unmatched reports $R_{j}$ ($j\neq i$)~\cite{CLIP2021}.
However, such a simplistic rule 
is not suitable for medical imaging scenarios, as reports from different medical images can often exhibit semantic consistency in terms of their interpretation.
We randomly sample 10 chest CT image-report pairs, and in Figure~\ref{fig:Pos_FN_pairs_vis}, visualize the semantic correspondence matrix between them. It can be observed that some images, in addition to their paired report, are semantically consistent with multiple other reports from different images. For example, the reports of the second, sixth, and ninth patients in the figure all show no abnormalities in the lung. This implies that the image of the second patient, $I_2$, not only matches with $R_2$ but also semantically aligns perfectly with $R_6$ and $R_9$.
%
Unfortunately, the traditional contrastive learning method would mistakenly consider these positive pairs, $R_6$ and $R_9$, as ``negative", leading to false negative pairs and sub-optimal optimization solutions.
To address this false negative issue, we propose the following robust contrastive loss:
\begin{equation}
\small
    L_{RoCo} = -\frac{1}{N *| P_i |}\sum_{i=1}^{N} \sum_{j\in P_i}\log(\frac{\exp(\left\langle I_i, T_j \right\rangle/t)}{\sum_{k=1}^N \exp(\left\langle I_i, T_k \right\rangle/t)}),
\end{equation}
where $\left\langle a,b \right\rangle$ refers to the cosine similarity of $a$ and $b$, and $P_i$ represents the positive text set for image $I_i$, which includes paired and semantically consistent reports. 
We define positive report pairs for an image using simple yet effective consistency rules, which involve determining whether two reports are identical. Many health reports have fixed content, making it easier to identify consistent pairs.
To address the issue of false negative pairs, we implement the following rules for correction. Firstly, regarding health reports, we identify all patients considered healthy by searching for key phrases like ``show no obvious abnormality" and ``show no active lesion" in the conclusion of the reports. Reports for these patients are deemed semantically matched. 
Secondly, in the case of abnormal reports, we stipulate that two reports must be identical in content to qualify as a semantic match.
This strict criterion is necessary because some reports may be 99\% similar in content, yet differ in 1\% crucial details such as the location or size of a lesion. These two rules ensure the precise identification of false negative pairs, thereby reducing the impact of noise samples on the quality of contrastive learning.

\subsubsection{Entity-focused masking (EFM)}
\label{sec.EFM}
In medical reports, the significance of words differs from that in natural language. For instance, in the sentence ``\textit{solid nodule in the right upper lung}", keywords of entities and attributes such as ``nodule', ``solid", ``right" and ``upper" hold more importance than common words like ``in" and ``the". This distinction is crucial for accurately diagnosing diseases. 
To emphasize this vital information, we introduce an EFM module, which is a specialized variant of masked language modeling. Diverging from BERT's approach of random masking~\cite{devlin2018bert}, the EFM selectively masks those entities and attributes with a higher probability. 
In addition, we need to be careful about information leakage when masking entities. For example, if we only mask ``nodule" in the phrase ``solid nodule", the language model can easily predict the next word as ``nodule" based on the context word ``solid" which hinders the model from extracting visual cues. Therefore, we intentionally treat the complete phrase ``solid nodule" as a whole masking object.
Notably, this encoder shares parameters with the multi-modal encoder of IRM.


\section{Experiments}

\subsection{Datasets}
\noindent\textbf{ChestCT-16K}: 
We collected a large dataset (named ChestCT-16K) of 3D chest CT volumes and corresponding radiology reports from a hospital, consisting of 16,685 consecutive patients. For test data integrity, we isolated the most recent month's data as our test set. From the remaining data, we randomly selected 10\% as the validation set. Thus, the dataset was split into 12,416 cases for training, 1,379 for validation, and 2,890 reserved for testing.

\noindent\textbf{ChestCT-EXT}: 
We additionally gathered 1,753 3D chest CT volumes and radiology reports from a separate hospital for external evaluation. It is noteworthy that ChestCT-EXT and ChestCT-16K are sourced from different hospitals.

\noindent\textbf{LIDC}: For the zero-shot task, a total of 876 CT images with nodules were utilized as an external test set~\cite{LIDC_dataset} to assess the model's generalization ability to detect nodules.


\subsection{Implementation details}
\noindent\textbf{Data pre-processing}: 
We use a lung segmentation model to crop the lung region of interest from the original CT images, and then directly resize it to a fixed size of (64, 224, 320). The Hounsfield Unit (HU) value is truncated to [-1150, 350] and subsequently normalized to [-1, 1].
\noindent\textbf{Pre-training}:
Each image token in ViT is processed with a patch size of (16, 32, 32), resulting in 280 image tokens per image. 
The batch size is set to 96, and the model undergoes training with 4 A100 GPUs for 200 epochs. The learning rate is gradually reduced from 1e-4 to 1e-6 following a cosine decay schedule.
\noindent\textbf{Report generation}: 
We commence this task by utilizing the whole BIUD framework, augmented with an added text decoder specifically for the generation of reports. This generation process is optimized by using Language Model (LM) loss. To enhance the efficiency, the text decoder is initialized with parameters derived from the pre-trained text encoder. 
\noindent\textbf{Zero-shot learning}:
We employ the pre-trained CT image encoder and text encoder for zero-shot classification, using prompts to facilitate this process. The default positive prompt template is structured as ``this is a chest CT with \{\} in lung" whereas the negative prompt template is phrased as ``this is a chest CT with no evident \{\} in lung". In these templates, \{\} serves as a placeholder for various disease entities. 
\noindent\textbf{Fine-tuning}:
The pre-trained CT image encoder serves as the backbone for various fine-tuning tasks. This is enhanced with modules tailored to specific tasks for the purpose of fine-tuning. Depending on the task, this may include integrating a classifier for classification tasks. 
\noindent\textbf{Evaluation metrics}:
We utilize the CheXpert labeling tool~\cite{irvin2019chexpert} to extract pathology entities. Unlike the original version, we add and modify some entities to fit our assessment scenario and set corresponding keywords based on the distribution of our data, following the advice of our collaborating doctors. The matching keywords for each entity extraction are provided in the appendix. We select the six most frequently occurring entities in the training data as our evaluation targets, including nodule, opacity, pleural effusion, emphysema, inflammation, and calcification.
Three metrics of precision, recall, and F1-score are utilized to assess the efficacy of different methods in zero-shot classification and report generation. Traditional NLP metrics like BLEU are not included in the main experiment results, as they may not accurately reflect clinical relevance in the reports. 

\subsection{Ablation study}

We validate the impact of three modules, EFM, RoCo, and KD, on the validation set using the report generation task through ablation studies. Our baseline model consists of three functional units: a basic contrastive learning unit, an image-report matching unit, and a language modeling unit.
As shown in Table~\ref{ablation_table}, each enhancement component contributes to the improvement of the model's performance, with the combination of EFM, ROCO, and KD providing the best overall F1 score. 
We also observe that the addition of KD shows an interesting shift in the metric of recall. While precision slightly decreases to 37.7\%, recall sees a significant increase to 40.8\%, leading to the highest F1 score in the table at 38.9\%. This significant increase in recall suggests that adding KD is particularly effective in identifying positive cases, which might be critical in applications where missing a positive case (such as a disease) has serious consequences.

\begin{table}[]
\small
\centering
\begin{tabular}{l|l|l|l}
\toprule
Methods          & P    & R    & F1   \\ \midrule
Baseline             & 35.6 & 34.6 & 35.0 \\ \hline
Baseline+EFM         & 37.8 & 35.9 & 36.7 \\ \hline
Baseline+RoCo        & 37.3 & 37.1 & 36.9 \\ \hline
Baseline+EFM+RoCo    & 38.8 & 36.6 & 37.5 \\ \hline
Baseline+EFM+RoCo+KD & 37.7 & 40.8 & 38.9 \\ \bottomrule
\end{tabular}
\vspace{-0.2cm}
\caption{The ablation study on three components of BIUD. EFM: entity-focused masking; RoCo: robust contrastive learning; KD: knowledge distillation.}
\vspace{-0.4cm}
\label{ablation_table}
\end{table}

\subsection{Report generation}
We compare three types of report generation methods. The first method involves pure language generation, which is trained directly on the ChestCT-16K training data using language modeling, such as R2Gen~\cite{R2Gen} and CMN~\cite{R2Gen-CMN}. 
The second method emphasizes contrastive pre-training, facilitating semantic alignment between image and report data prior to the generation of language, exemplified by methods like BLIP~\cite{BLIP} and DCL~\cite{DCL}.
The third approach we assessed involves RadFM~\cite{RadFM}, a large foundational model that has undergone extensive pre-training across a diverse array of multimodal tasks. This model is notable for its direct application to a variety of tasks without the necessity for additional training. Significantly, RadFM has also been trained using a dataset comprising chest CT images and accompanying reports.

We first compare these different methods on our internal ChestCT-16K test data, as listed in Table~\ref{tab.report_generation}. Based on this comparison, we draw three key conclusions: 
First, methods that directly generate reports are underperformed by those that employ image-report contrastive pre-training, proving that aligning images and reports in semantic space is beneficial for language generation. 
Second, although RadFM has the advantage of task diversity and can handle many different types of tasks, its generalization for specific diagnostic tasks still has considerable room for improvement. For example, it almost lacks the detection capability (very low recall) for some basic diseases on our test set. 
Lastly, our method shows the best overall performance on six disease diagnostic tasks compared to other methods, achieving precision, recall, and F1-scores of 36.2, 38.9, and 37.3, respectively. 
To confirm its robustness, we further conduct external testing on the ChestCT-EXT data, where our model still maintained good generalizability, especially surpassing the internal test performance on Nodule and Opacity tasks. The overall highest F1 scores of our method demonstrate a better balance between precision and recall.

\begin{table*}[]
\small
\centering
\scalebox{0.78}{
\begin{tabular}{lccccccccccccccccccccc}
\toprule
\multicolumn{1}{c|}{\multirow{2}{*}{Methods}} & \multicolumn{3}{c|}{Nodule}             & \multicolumn{3}{c|}{Opacity}            & \multicolumn{3}{c|}{Pleural Effusion}   & \multicolumn{3}{c|}{Emphysema}          & \multicolumn{3}{c|}{Inflammation}       & \multicolumn{3}{c|}{Calcification}      & \multicolumn{3}{c}{Average} \\
\multicolumn{1}{c|}{}                         & P    & R    & \multicolumn{1}{c|}{F1}   & P    & R    & \multicolumn{1}{c|}{F1}   & P    & R    & \multicolumn{1}{c|}{F1}   & P    & R    & \multicolumn{1}{c|}{F1}   & P    & R    & \multicolumn{1}{c|}{F1}   & P    & R    & \multicolumn{1}{c|}{F1}   & P       & R       & F1      \\ \hline
\multicolumn{22}{c}{Internal test: ChestCT-16K}                                                                                                                                                                                                                                                                                                       \\ \midrule
\multicolumn{1}{l|}{R2Gen~\cite{R2Gen}}                    & 70.6 & 17.4 & \multicolumn{1}{c|}{27.9} & 65.1 & 10.0 & \multicolumn{1}{c|}{17.4} & 34.8 & 20.0 & \multicolumn{1}{c|}{25.4} & 0    & 0    & \multicolumn{1}{c|}{0}    & 20.4 & 3.3  & \multicolumn{1}{c|}{5.7}  & 20.6 & 5.7  & \multicolumn{1}{c|}{9.0}  & 35.3    & 9.4     & 14.2    \\
\multicolumn{1}{l|}{CMN~\cite{R2Gen-CMN}}                      & 73.2 & 7.5  & \multicolumn{1}{c|}{13.6} & 76.2 & 11.6 & \multicolumn{1}{c|}{20.2} & 43.4 & 27.5 & \multicolumn{1}{c|}{33.7} & 29.3 & 17.9 & \multicolumn{1}{c|}{22.2} & 34.3 & 7.9  & \multicolumn{1}{c|}{12.9} & 19.5 & 3.9  & \multicolumn{1}{c|}{6.5}  & 46      & 12.7    & 18.2    \\
\multicolumn{1}{l|}{RadFM~\cite{RadFM}}                    & 60.0 & 2.2  & \multicolumn{1}{c|}{4.3}  & 53.0 & 13.2 & \multicolumn{1}{c|}{21.2} & 33.3 & 0.83 & \multicolumn{1}{c|}{1.6}  & 0    & 0    & \multicolumn{1}{c|}{0}    & 0    & 0    & \multicolumn{1}{c|}{0}    & 0    & 0    & \multicolumn{1}{c|}{0}    & 24.4    & 2.7     & 4.5     \\
\multicolumn{1}{l|}{BLIP~\cite{BLIP}}                     & 66.2 & 65.1 & \multicolumn{1}{c|}{65.6} & 55.5 & 54.9 & \multicolumn{1}{c|}{55.2} & 30.1 & 30.8 & \multicolumn{1}{c|}{30.5} & 19.8 & 24.2 & \multicolumn{1}{c|}{21.8} & 26.7 & 18.5 & \multicolumn{1}{c|}{21.8} & 15   & 20.6 & \multicolumn{1}{c|}{17.3} & 35.6    & 35.7    & 35.4    \\
\multicolumn{1}{l|}{DCL~\cite{DCL}}                      & 65.7 & 62.8 & \multicolumn{1}{c|}{64.3} & 56.0 & 53.9 & \multicolumn{1}{c|}{54.9} & 29.1 & 32.5 & \multicolumn{1}{c|}{30.7} & 20.8 & 26.3 & \multicolumn{1}{c|}{23.3} & 24.1 & 19.8 & \multicolumn{1}{c|}{21.7} & 17.8 & 20.8 & \multicolumn{1}{c|}{19.2} & 35.6    & 36.0    & 35.7    \\ \midrule
\multicolumn{1}{l|}{Our BIUD}                     & 65.4 & 76.3 & \multicolumn{1}{c|}{70.4} & 56.3 & 59.4 & \multicolumn{1}{c|}{57.8} & 30.7 & 31.7 & \multicolumn{1}{c|}{31.2} & 23.9 & 28.4 & \multicolumn{1}{c|}{26.0} & 26.3 & 18.5 & \multicolumn{1}{c|}{21.7} & 14.7 & 19.3 & \multicolumn{1}{c|}{16.7} & 36.2    & 38.9    & 37.3    \\ \bottomrule
\multicolumn{22}{c}{External test: ChestCT-EXT}                                                                                                                                                                                                                                                                                                       \\ \midrule
\multicolumn{1}{l|}{R2Gen~\cite{R2Gen}}                    & 85.5 & 43.0 & \multicolumn{1}{c|}{57.2} & 67.5 & 20.9 & \multicolumn{1}{c|}{31.9} & 16.5 & 15.8 & \multicolumn{1}{c|}{16.1} & 20.0 & 0.7  & \multicolumn{1}{c|}{1.3}  & 15.6 & 5.51 & \multicolumn{1}{c|}{8.1}  & 12.8 & 9.4  & \multicolumn{1}{c|}{10.8} & 36.3    & 15.9    & 20.9    \\
\multicolumn{1}{l|}{CMN~\cite{R2Gen-CMN}}                      & 78.0 & 16.0 & \multicolumn{1}{c|}{26.5} & 67.9 & 21.1 & \multicolumn{1}{c|}{32.1} & 26.8 & 27.4 & \multicolumn{1}{c|}{27.1} & 36.2 & 11.0 & \multicolumn{1}{c|}{16.8} & 21.8 & 7.5  & \multicolumn{1}{c|}{11.1} & 16.0 & 8.5  & \multicolumn{1}{c|}{11.1} & 41.1    & 15.3    & 20.9    \\
\multicolumn{1}{l|}{RadFM~\cite{RadFM}}                    & 79.5 & 5.3  & \multicolumn{1}{c|}{9.9}  & 67.5 & 9.7  & \multicolumn{1}{c|}{17.0} & 100  & 0.7  & \multicolumn{1}{c|}{1.4}  & 0    & 0    & \multicolumn{1}{c|}{0}    & 0    & 0    & \multicolumn{1}{c|}{0}    & 0    & 0    & \multicolumn{1}{c|}{0}    & 41.2    & 2.6     & 4.7     \\
\multicolumn{1}{l|}{BLIP~\cite{BLIP}}                     & 85.7 & 75.6 & \multicolumn{1}{c|}{80.3} & 64.8 & 66.2 & \multicolumn{1}{c|}{65.5} & 19.7 & 39.0 & \multicolumn{1}{c|}{26.2} & 25.0 & 23.9 & \multicolumn{1}{c|}{24.4} & 13.0 & 13.4 & \multicolumn{1}{c|}{13.2} & 11.4 & 20.1 & \multicolumn{1}{c|}{14.6} & 36.6    & 39.7    & 37.4    \\
\multicolumn{1}{l|}{DCL~\cite{DCL}}                      & 85.0 & 73.0 & \multicolumn{1}{c|}{78.5} & 67.0 & 68.8 & \multicolumn{1}{c|}{67.9} & 18.5 & 40.0 & \multicolumn{1}{c|}{25.3} & 21.5 & 20.0 & \multicolumn{1}{c|}{20.7} & 18.2 & 23.2 & \multicolumn{1}{c|}{20.4} & 12.0 & 21.0 & \multicolumn{1}{c|}{15.2} & 37.0    & 41.0    & 38.0    \\ \midrule
\multicolumn{1}{l|}{Our BIUD}                     & 86.7 & 83.9 & \multicolumn{1}{c|}{85.3} & 67.2 & 73.6 & \multicolumn{1}{c|}{70.3} & 21.4 & 43.2 & \multicolumn{1}{c|}{28.6} & 25.2 & 25.2 & \multicolumn{1}{c|}{25.2} & 16.4 & 17.7 & \multicolumn{1}{c|}{17.0} & 13.2 & 24.6 & \multicolumn{1}{c|}{17.1} & 38.4    & 44.7    & 40.6    \\ \bottomrule
\end{tabular}
}
\vspace{-0.2cm}
\caption{Report generation results evaluated by disease diagnosis on the ChestCT-16K test set and ChestCT-EXT dataset. P: precision; R: recall; F1: F1-score.}
\vspace{-0.4cm}
\label{tab.report_generation}
\end{table*}

\subsection{Zero-shot diagnosis}

We first evaluate the zero-shot capabilities of various models on the ChestCT-16K test set. Table~\ref{tab.zeroshot} presents the average performance of different methods across six diagnostic tasks. Our method shows a substantially higher recall rate of 57.7\% compared to other methods and achieves the highest F1 score of 33.0\%, indicating an overall superior performance in accurately identifying relevant cases.
Additionally, we evaluate the model's capacity for detecting pulmonary nodules on an external dataset, LIDC. Our method demonstrated a markedly superior ability to detect nodules compared to the other methods. 

\begin{figure}
    \centering
    \includegraphics[width=1\linewidth]{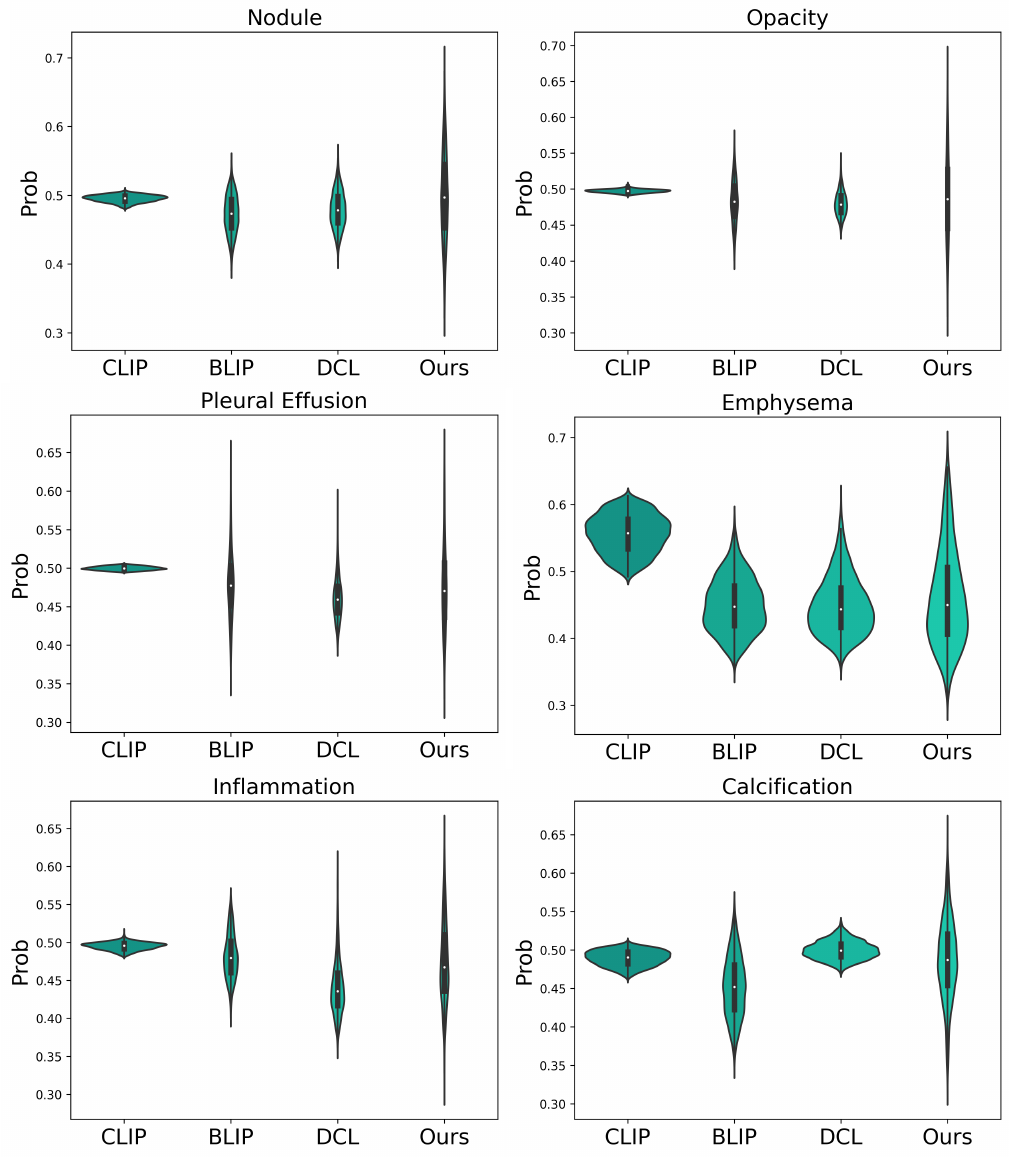}
    \vspace{-0.4cm}
    \caption{Violin plots illustrating the distribution of zero-shot classification probabilities obtained by four models, CLIP, BLIP, DCL, and our proposed BIUD, across six tasks.}
    \vspace{-0.2cm}
    \label{fig:violin}
\end{figure}

We provide the violin plots of the probability distribution for each model during zero-shot classification, illustrated in Figure~\ref{fig:violin}. 
The models compared, such as CLIP, exhibit probability distributions that gravitate around the midpoint of 0.5 for the majority of tasks. This reflects a slight edge in the correct classification of certain samples. However, it is accompanied by a notably low level of confidence, which means the embedding between positive and negative samples is relatively indistinct. Conversely, our model demonstrates a wider dispersion in probability distributions, with a clear deviation from the central value of 0.5, which denotes a heightened level of confidence in its decision-making process. 
Besides, there is a clear imbalance in probability distributions across various tasks in the compared models. For example, the average probability of DCL falls below 0.5 in tasks such as ``\textit{Inflammation}", implying a tendency to fail to detect a majority of inflammation-positive samples. In stark contrast, our model maintains a more equitable probability distribution throughout most tasks, which is instrumental in enabling our model to maintain an optimal balance between precision and recall.
These observations are indicative of our model's superior capability in aligning images with their corresponding reports.

\begin{table}[]
\small
\centering
\scalebox{0.9}{
\begin{tabular}{l|ccc|ccc}
\toprule
\multicolumn{1}{c|}{\multirow{2}{*}{Methods}} & \multicolumn{3}{c|}{Internal test: ChestCT-16K} & \multicolumn{3}{c}{External test: LIDC} \\
\multicolumn{1}{c|}{}                         & P           & R          & F1         & P       & R      & F1     \\ \midrule
CLIP~\cite{CLIP2021}                                          & 26.5        & 30.7       & 16.0       & 84.9    & 9.7    & 17.4   \\
BLIP~\cite{BLIP}                                          & 26.6        & 31.9       & 21.6       & 91.7    & 4.4    & 8.4    \\
DCL~\cite{DCL}                                           & 32.9        & 35.6       & 25.7       & 86.9    & 41.4   & 56.0   \\ \midrule
Our BIUD                                          & 30.6        & 57.7       & 33.0       & 83.9    & 54.1   & 65.8   \\ \bottomrule
\end{tabular}
}
\vspace{-0.2cm}
\caption{Zero-shot classification results evaluated by disease diagnosis on ChestCT-16K test set and LIDC dataset.}
\label{tab.zeroshot}
\end{table}

\subsection{Fine-tuning}
We conduct a fine-tuning experiment to further validate the effectiveness of our pre-trained model, BIUD. Utilizing the modified CheXpert labeling tool, we extract labels for 6 entities on ChestCT-16K, where 0 indicates the absence and 1 indicates the presence of a symptom.
We employ the CT image encoder for this multi-label classification task. On the ChestCT-16K test data, we compare different pre-trained models, including those pre-trained on ImageNet (IN) and our BIUD.
The results in Table~\ref{tab.fine-tuning} show that using BIUD for fine-tuning significantly enhances the ability to detect various entities, albeit with a slight decrease in precision. Maintaining a balance between precision and recall, our method improves the F1-score by 4.5\% compared to the ImageNet fine-tuning.
Another interesting finding is that models initialized with ImageNet pre-training exhibit slightly lower performance compared to the results in report generation (Table \ref{tab.report_generation}), with an F1-score of 37.3 vs. 34.5. The key difference between these two approaches is that one uses language as the supervision signals, while the other uses 0-1 categorical labels. We believe that the density of information provided in language far exceeds that in one-hot labels, which is invaluable for analyzing 3D medical images of complex anatomical structures. On one hand, it can prevent the model from overfitting, and on the other hand, it can help the model uncover more associative information to better understand images. For instance, pleural effusion often accompanies blurred shadows at the lung margins, which can provide additional information for decision-making.
These results reveal the enormous potential of language-driven image understanding in medical imaging.

\begin{table}[]
\small
\centering
\scalebox{0.9}{
\begin{tabular}{l|cc|cc|cc}
\toprule
\multirow{2}{*}{Pythology} & \multicolumn{2}{c|}{P} & \multicolumn{2}{c|}{R} & \multicolumn{2}{c}{F1} \\
                           & IN         & BIUD      & IN         & BIUD      & IN         & BIUD      \\ \midrule
Nodule                     & 66.3       & 65.3      & 76.4       & 76.3      & 71.0       & 70.4      \\
Opacity                    & 57.6       & 54.2      & 55.1       & 74.4      & 56.3       & 62.7      \\
Pleural Effusion           & 54.7       & 65.5      & 29.2       & 30.0      & 38.0       & 41.1      \\
Emphysema                  & 34.9       & 37.5      & 15.8       & 15.8      & 21.7       & 22.2      \\
Inflammation               & 38.8       & 31.3      & 10.2       & 20.5      & 16.2       & 24.8      \\
Calcification              & 27.6       & 18.9      & 2.1        & 9.6       & 3.9        & 12.8      \\ \midrule
Average                    & 46.7       & 45.5      & 31.5       & 37.8      & 34.5       & 39.0      \\ \bottomrule
\end{tabular}
}
\vspace{-0.2cm}
\caption{Comparison of fine-tuning results of ImageNet pre-training model (IN) and our BIUD pre-training model on the ChestCT-16K test set.}
\vspace{-0.4cm}
\label{tab.fine-tuning}
\end{table}

\subsection{Comparison to radiologist}
\begin{figure}
    \centering
    \includegraphics[width=1.0\linewidth]{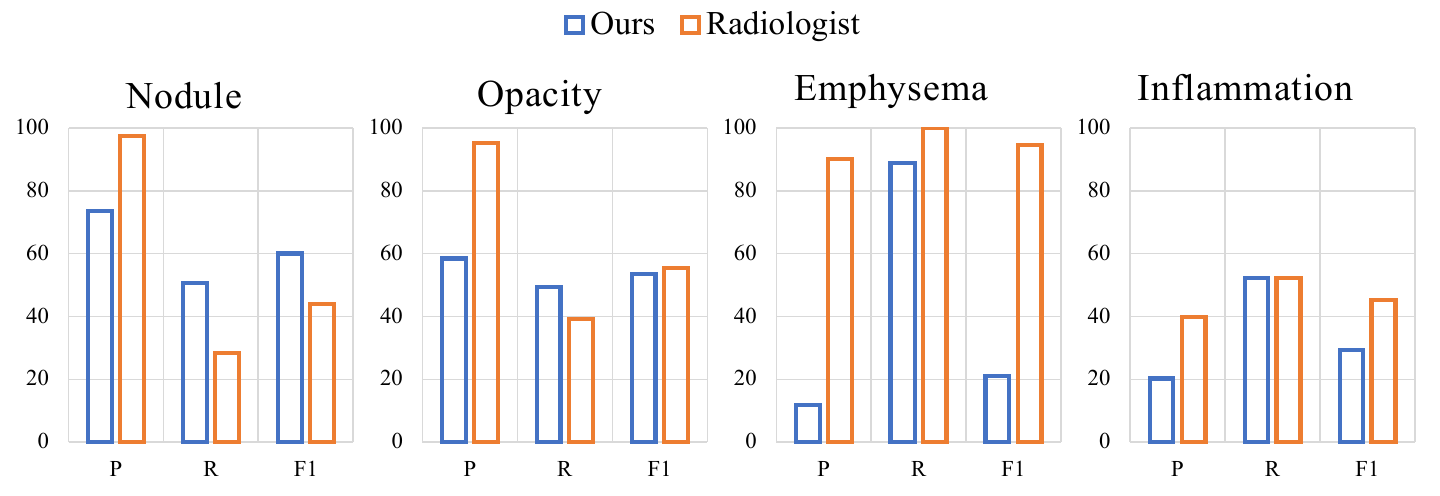}
    \vspace{-0.7cm}
    \caption{Comparison to a radiologist having over 10 years of experience across four tasks.}
    \vspace{-0.3cm}
    \label{fig:reader_study}
\end{figure}

We randomly selected 200 patients from the ChestCT-16K test set for a reader study. A radiologist with over 10 years of experience conducted image assessments in a continuous session. Without any prior information, the radiologist was tasked with identifying all possible abnormalities from these test images.
Using actual reports as a reference, we compared the radiologist’s assessments with our method in Figure~\ref{fig:reader_study}. It's important to note that, in real clinical practice, completing an actual report typically takes about 3 to 4 minutes, while in our assessment, the radiologist aimed to expedite the process, spending on average about 30 seconds per case. In contrast, our model processed each case in less than 1 second. The radiologist’s diagnostic performance in certain tasks, such as nodule and pleural effusion detection, declined under the high-intensity work environment, possibly due to reduced concentration, underscoring the necessity for developing automated AI-assisted diagnostic tools.
The results also revealed that the radiologist’s precision was higher than our model, but our model had an advantage in recall for most tasks. This means that there were instances of missed diagnoses by the radiologist. In comparison, our model could perform repetitive tasks with unbiased consistency. Overall, our model demonstrates promising capabilities in understanding medical images, achieving a performance comparable to that of a radiologist in detecting nodules. These results highlight the potential of language-driven medical image understanding and its promising prospects for revolutionizing clinical diagnostics in the future.

\section{Conclusion and discussion}

In this paper, we initially explore the feasibility of enhancing the understanding of medical images using reports, especially in 3D CT scenarios lacking sufficient paired data. 
We propose distilling knowledge from the 2D X-ray expert model to bootstrap the semantic representation of 3D CT models. To align the two modalities, we introduce a robust contrastive learning approach to minimize the impact of false negative pairs on semantic alignment. 
Our model demonstrates superior performance across various tasks, including zero-shot learning, report generation, and fine-tuning. Remarkably, it even matches the diagnostic abilities of a doctor with over 10 years of experience in detecting lung nodules.
This research initially unveils the tremendous potential of reports in aiding medical image understanding, potentially inspiring new directions in the development of annotation-free medical AI models. While some promising results have been achieved in 3D medical imaging, current language-driven image understanding models still have a long way to go before clinical application. Their generalization capabilities and application boundaries remain to be explored and validated in the future.

\noindent \textbf{Acknowledgements.} W. Cao and J. Zheng were supported by the National Natural Science Foundation of China under Grants 62371449 and U23A20483.
\newpage
{
    \small
    \bibliographystyle{ieeenat_fullname}
    \bibliography{main}
}

\clearpage
\setcounter{page}{1}
\maketitlesupplementary

\section{Extraction of pathology entities}
We utilized the CheXpert labeling tool~\cite{irvin2019chexpert} to identify six pathologies within the reports. For each CT report, specific keywords were used to pinpoint six different pathologies, detailed in Table~\ref{tab.key_words}.
The frequency of each pathology in the ChestCT-16K and ChestCT-EXT datasets is summarized in Table~\ref{tab.pathology_number}.

\begin{table}[H]
\small
\centering
\begin{tabular}{l|clllll}
\hline
Pythology        & \multicolumn{6}{c}{Related key words}                                                                                                                                                                                                                                                                                                                                                                                             \\ \hline
Nodule           & \multicolumn{6}{c}{nodule, nodules, nodular}                                                                                                                                                                                                                                                                                                                                                                                      \\ \hline
Opacity          & \multicolumn{6}{c}{\begin{tabular}[c]{@{}c@{}}opacity, opacities, decreased  translucency, \\ increased density,  airspace disease, \\ air-space disease, air space \\ disease, infiltrate, infiltration, \\ interstitial marking,  interstitial pattern, \\ interstitial lung, reticular pattern, \\ reticular marking, reticulation, \\ parenchymal scarring,  peribronchial \\ thickening, wall thickening, scar\end{tabular}} \\ \hline
Pleural Effusion & \multicolumn{6}{c}{pleural   fluid, pleural effusion}                                                                                                                                                                                                                                                                                                                                                                             \\ \hline
Emphysema        & \multicolumn{6}{c}{emphysema}                                                                                                                                                                                                                                                                                                                                                                                                     \\ \hline
Inflammation     & \multicolumn{6}{c}{\begin{tabular}[c]{@{}c@{}}inflammation, pneumonia, \\      infection, infectious \\      process, infectious\end{tabular}}                                                                                                                                                                                                                                                                                    \\ \hline
Calcification    & \multicolumn{6}{c}{calcification,   calcifications}                                                                                                                                                                                                                                                                                                                                                                               \\ \hline
\end{tabular}
\caption{Keywords for extracting six pathologies.}
\label{tab.key_words}
\end{table}

\begin{table}[H]
\small
\centering
\begin{tabular}{l|ccc|c}
\toprule
\multirow{2}{*}{Pythology} & \multicolumn{3}{c|}{ChestCT-16K} & ChestCT-EXT \\
                           & Train      & Val      & Test     & Test        \\ \midrule
Nodule                     & 8084       & 923      & 1894     & 1689        \\
Opacity                    & 6559       & 757      & 1489     & 1347        \\
Pleural Effusion           & 536        & 65       & 120      & 147         \\
Emphysema                  & 522        & 57       & 95       & 221         \\
Inflammation               & 1244       & 139      & 303      & 220         \\
Calcification              & 1826       & 182      & 384      & 290         \\ \bottomrule
\end{tabular}
\caption{The number of six entities in the ChestCT-16K and ChestCT-EXT datasets.}
\label{tab.pathology_number}
\end{table}

\section{Comparison of different report generation methods using NLP metrics}
Table~\ref{tab.nlp_results} shows the NLP evaluation of our method, consistently outperforming other competitors across all NLP metrics, not only on the internal test set, ChestCT-16K, but also on the external test set, ChestCT-EXT.

\begin{table}[]
\small
\centering
\begin{tabular}{lcccc}
\toprule
\multicolumn{1}{l|}{Methods} & BLEU-4 & METEOR & ROUGE-L & CIDEr \\ \hline
\multicolumn{5}{c}{Internal test: ChestCT-16K}                   \\ \toprule
\multicolumn{1}{l|}{R2Gen~\cite{R2Gen}}   & 7.3    & 10.4   & 28.3    & 7.8   \\
\multicolumn{1}{l|}{CMN~\cite{R2Gen-CMN}}     & 5.6    & 8.6    & 25.5    & 5.2   \\
\multicolumn{1}{l|}{RadFM~\cite{RadFM}}   & 0.1    & 4.3    & 8.6     & 1.5   \\
\multicolumn{1}{l|}{BLIP~\cite{BLIP}}    & 13.2   & 19.4   & 37.9    & 21.1  \\
\multicolumn{1}{l|}{DCL~\cite{DCL}}     & 12.5   & 18.6   & 37.1    & 19.0  \\ \midrule
\multicolumn{1}{l|}{Ours}    & \textbf{13.8}   & \textbf{21.3}   & \textbf{40.2}    & \textbf{24.4}  \\ \bottomrule
\multicolumn{5}{c}{External test: ChestCT-EXT}                   \\ \toprule
\multicolumn{1}{l|}{R2Gen~\cite{R2Gen}}   & 6.9    & 10.8   & 27.1    & 3.7   \\
\multicolumn{1}{l|}{CMN~\cite{R2Gen-CMN}}     & 3.2    & 7.2    & 21.3    & 1.7   \\
\multicolumn{1}{l|}{RadFM~\cite{RadFM}}   & 0.1    & 0.7    & 1.6     & 0.2   \\
\multicolumn{1}{l|}{BLIP~\cite{BLIP}}    & 11.2   & 16.7   & 33.8    & 9.6   \\
\multicolumn{1}{l|}{DCL~\cite{DCL}}     & 10.6   & 16.4   & 33.6    & 9.9   \\ \midrule
\multicolumn{1}{l|}{Ours}    & \textbf{12.0}   & \textbf{17.6}   & \textbf{35.4}    & \textbf{10.7}  \\ \bottomrule
\end{tabular}
\caption{Comparison of NLP metrics of different methods in the ChestCT-16K test set and ChestCT-EXT dataset.}
\label{tab.nlp_results}
\end{table}

\begin{figure*}
    \centering
    \includegraphics[width=1.0\linewidth]{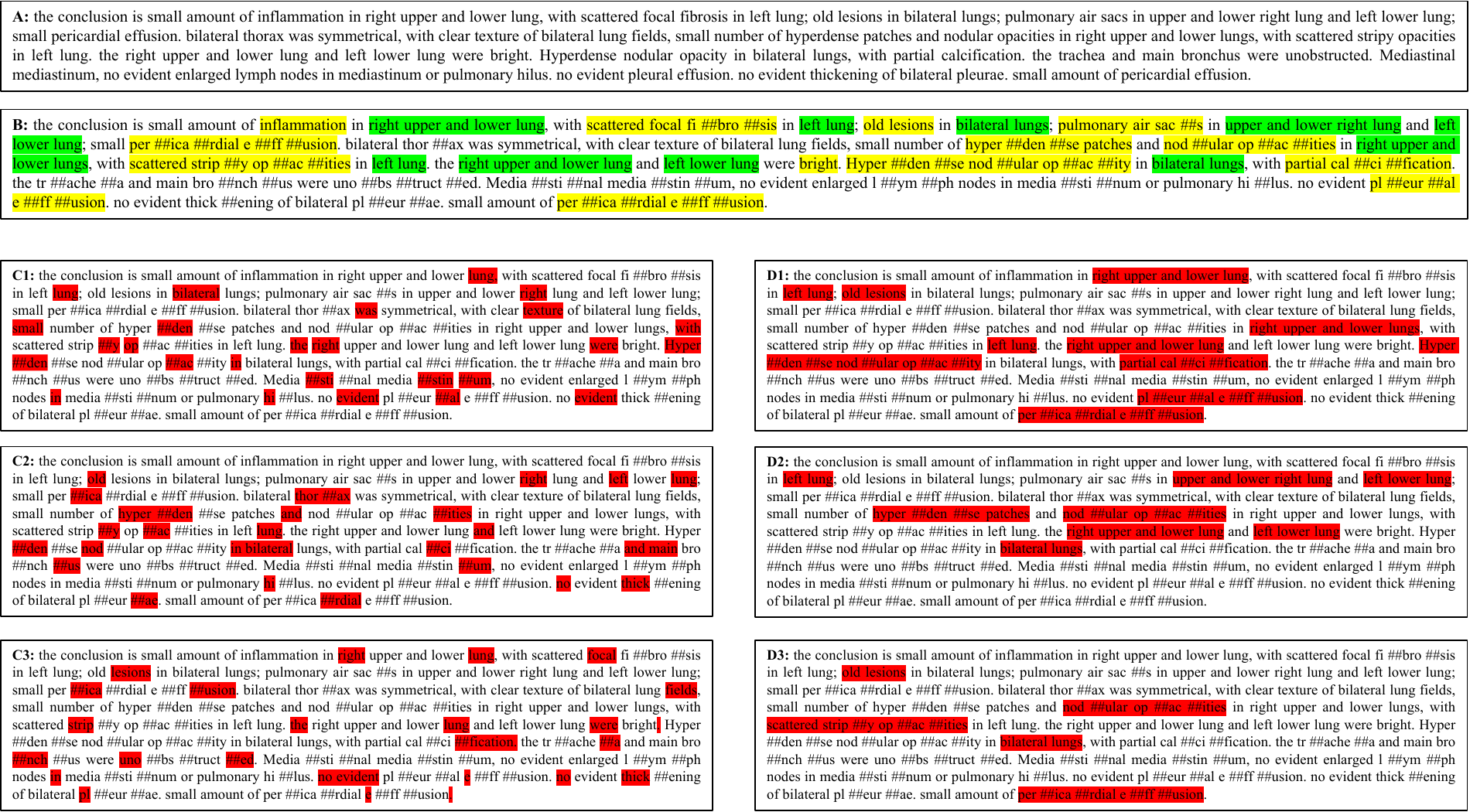}
    \caption{Comparison of random masking and our proposed entity-focused masking.}
    \label{fig:masking}
\end{figure*}

\section{Comparison of entity-focused masking (EFM) and random masking}
Figure~\ref{fig:masking} illustrates random masking and the proposed EFM. The panel `A' displays the original report, while the panel `B' illustrates the tokenized report using the BERT tokenizer~\cite{devlin2018bert}, with the `\#\#' symbol indicating the sub-word of the previous one. 
In the panel B, the highlighted entities are marked in yellow, and the attributes of these entities, such as location, are highlighted in green. The panels `C1', `C2', and `C3' demonstrate examples of the report with tokens randomly masked (indicated by red highlights), compared to the panels `D1', `D2', and `D3', which utilize the EFM for selective masking of reports. 
The illustration makes it clear that the random masking may obscure words like `lung' or `thick' that could be guessed from the surrounding text, whereas EFM masks either an entity or its attributes. This encourages the model to rely more heavily on visual clues from the associated images to predict the masked information.

\section{An example of language-guided retrieval}

\begin{figure*}
    \centering
    \includegraphics[width=1.0\linewidth]{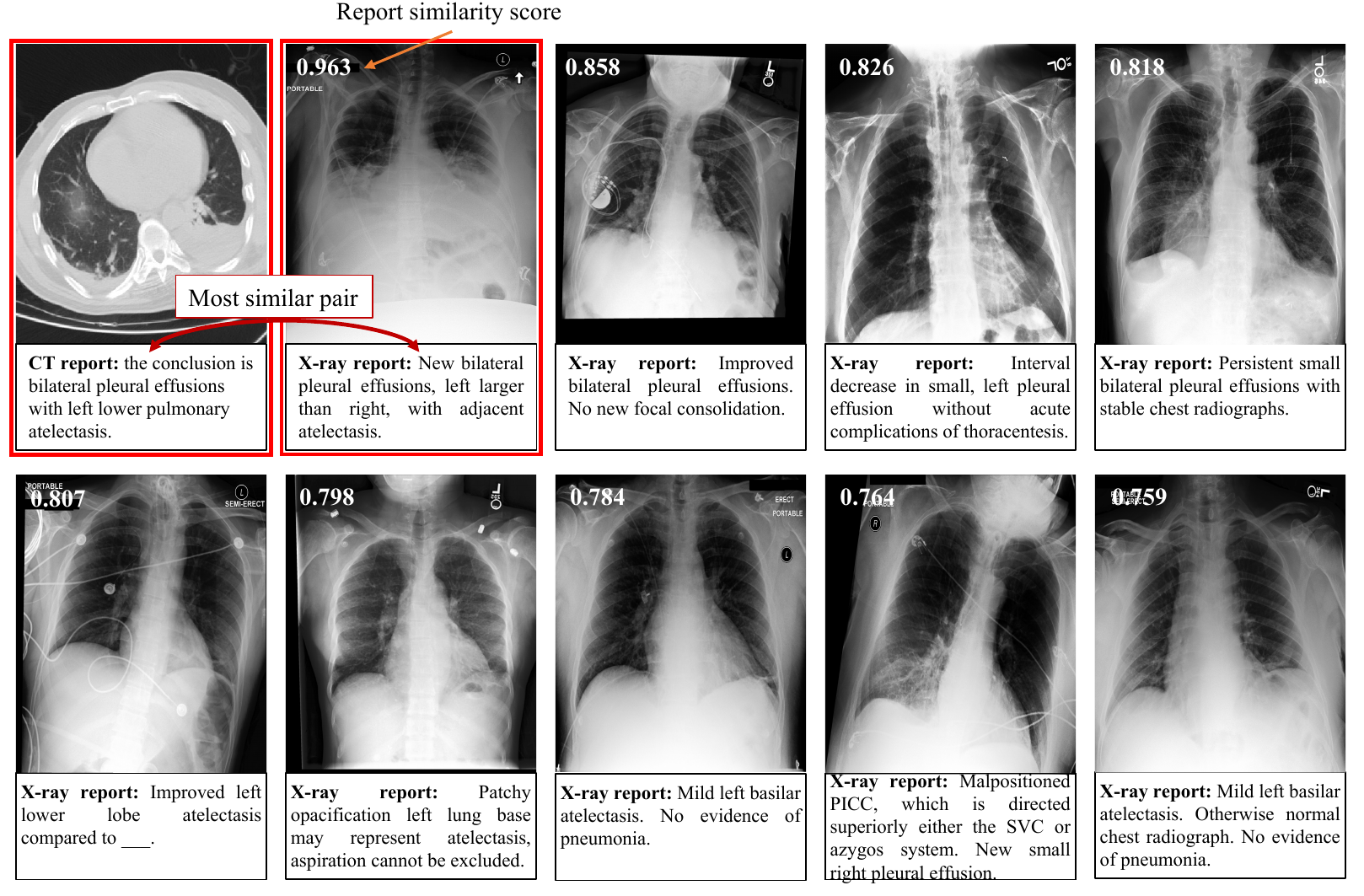}
    \caption{An example of language-guided report retrieval.}
    \label{fig:retrival}
\end{figure*}

In Figure~\ref{fig:retrival}, we present a detailed example of a language-guided retrieval process, which includes a CT query report and the nine most similar X-ray reports. The X-ray report with the highest degree of similarity has a score of 0.963. Symptoms of pleural effusion are present in both of the matching images. This retrieval method, which is based on the similarity of reports, can effectively identify the X-ray image that best matches each CT image query.

\end{document}